%% file: main.tex
\RequirePackage{fix-cm}
\documentclass[smallextended]{svjour3}       % onecolumn (second format)
\smartqed  % flush right qed marks, e.g. at end of proof
\usepackage{graphicx}
\usepackage{natbib}
\usepackage{xcolor}
\usepackage{amsmath}
\usepackage{amsfonts}
\def\makeheadbox{\relax}

\title{PARIS: Personalized Activity Recommendation for Improving Sleep Quality%using Individual Health and Lifestyle Constraints
}
\titlerunning{PARIS: Personalized Activity Recommendation for Improving Sleep Quality}        % if too long for running head

\author{Meghna Singh \and
        Saksham Goel \and
        Abhiraj Mohan \and
        Jaideep Srivastava %etc.
}

\authorrunning{Meghna Singh, 
        Saksham Goel,
        Abhiraj Mohan et al.}
\institute{
           Meghna Singh, Saksham Goel, Abhiraj Mohan, Jaideep Srivastava \at
               University of Minnesota, MN, USA \\
              \email{\{singh742, goelx29, mohan056, srivasta\}@umn.edu}
              \and
}

\date{}
\begin{document}
\maketitle

\begin{abstract}
\input{0abstract.tex}

\end{abstract}

\section{Introduction}
\input{1introduction.tex}
%merge introduction +background
%\section{Background}
%\input{2background.tex}

\section{Problem Definition}
\input{2problem_definition.tex}

\section{Solution Approach}
\input{3methodology.tex}

%merge this within experiments and results
%\section{Data}
%\input{3preliminaries.tex}

\section{Experiments and Results}
\input{4expts_results.tex}
%impact and discussion in this sectoion?

\section{Discussion}
\input{5discussion.tex}

%short conclusion and future work
\section{Conclusion}
\input{6conclusion.tex}

\begin{acknowledgements}
The Hispanic Community Health Study/Study of Latinos (HCHS/SOL) was performed as a collaborative study supported by contracts from the NHLBI to the University of North Carolina (N01-HC65233), University of Miami (N01-HC65234), Albert Einstein College of Medicine (N01-HC65235), Northwestern University (N01-HC65236), and San Diego State University (N01-HC65237) (AG05407, AR35582, AG05394, AR35584, AR35583, AG08415). The National Sleep Research Resource was supported by the National Heart, Lung, and Blood Institute (R24 HL114473, RFP 75N92019R002).
\end{acknowledgements}

\bibliographystyle{spbasic}      % basic style, author-year citations
\bibliography{main.bib}   % name your BibTeX data base

\end{document}

%% file: 0abstract.tex
The quality of sleep has a deep impact on people's physical and mental health. People with insufficient sleep are more likely to report physical and mental distress, activity limitation, anxiety, and pain. Moreover, in the past few years, there has been an explosion of applications and devices for activity monitoring and health tracking. Signals collected from these wearable devices can be used to study and improve sleep quality. 
In this paper, we utilize the relationship between physical activity and sleep quality to find ways of assisting people improve their sleep using machine learning techniques. %Heart rate along with body metrics like age, gender, and body mass index (BMI) are used to group biologically similar people into clusters. 
People usually have several behavior modes that their bio-functions can be divided into. %TODO [REF]. 
%The intraday heart rate and step count patterns reflect behavior modes that subjects' bio-functions can be divided into. 
Performing time series clustering on activity data, we find cluster centers that would correlate to the most evident behavior modes for a specific subject. Activity recipes are then generated for good sleep quality for each behavior mode within each cluster. 
These activity recipes are supplied to an activity recommendation engine for suggesting a mix of relaxed to intense activities to subjects during their daily routines. The recommendations are further personalized based on the subjects' lifestyle constraints, i.e. their age, gender, body mass index (BMI), resting heart rate, etc., with the objective of the recommendation being the improvement of that night's quality of sleep. This would in turn serve a longer-term health objective, like lowering heart rate, improving the overall quality of sleep, etc.

%% file: 1introduction.tex
%Sleep is vital for health. 
People spend a third of their lives sleeping because it is one of the most vital activities for maintaining good health. 
%Proper sleep not only increases the physical and mental alertness and productivity of people, it also reduces the risk for diseases like stroke and diabetes. \citep{taheri06,palotti2019benchmark}
%People spend a third of their lives asleep, and poor sleep quality has been linked to various chronic health conditions like obesity and diabetes \citep{taheri06,palotti2019benchmark}. There are many theories concerning what leads to improved sleep such as differences in activity level, afternoon rests, time of sleep etc.
The quality of sleep has a deep impact on people's physical and mental health. Poor sleep quality has been linked to various chronic health conditions like obesity and diabetes \citep{taheri06,knutson2006role, palotti2019benchmark} as well as cardiovascular diseases ~\citep{kasasbeh2006inflammatory} and depression~\citep{murphy2015sleep}.
%A good sleep can keep a person active and fresh whereas a poor sleep can lead to drowsiness, headache, and lethargy. ~\citep{strine2005associations} show that insufficient sleep is related to health related quality of life and health behaviors. 
%People with insufficient sleep are more likely to report physical and mental distress, activity limitation, anxiety and pain. Decreased sleep duration or quality may increase diabetes risk ~\citep{knutson2006role}, severe cardiovascular consequences~\citep{kasasbeh2006inflammatory}, depression~\citep{murphy2015sleep}, etc. 
%A study~\cite{williamson2000moderate} shows that moderate deprivation of sleep affects cognitive and motor performance of a person in the same way as alcohol does. Because of the great impact of sleep in people's overall well-being, sleep science has become an active research area in recent years.
Two recent studies \citep{blume2020effects, wright2020sleep} have shown how movement restrictions imposed due to COVID-19 led to an increase in sleep duration, but lowering of the quality of sleep because of increase in perceived burden. 

%Because of the impact of sleep in public health, and the increase in number of people with sleep related diseases, large companies like Apple, Google, Samsung, Fitbit along with start-ups are investing in sleep technology and sleep related research. 
%Polysomnography (PSG), a test to diagnose sleep disorder, was developed in 1950s~\citep{hirshkowitz2015history}. It records brain waves, the oxygen level in blood, heart rate and breathing, as well as eye and leg movements during the test.
%PSG measures the brain activity, eye movements, muscle activity, and heart rhythm. 
%But the test has to be done in a sleep lab or hospital. Since PSG is a high-fidelity test, it is considered as a gold standard for sleep research. But PSG is not scalable, it is too expensive and it does not consider physical activity during the day.
Polysomnography (PSG) is the state-of-the-art and most accurate method to measure sleep quality. PSG measures brain activity, eye movements, muscle activity, and heart rhythm. But it is a highly intrusive and costly approach. Due to its complexity, it is typically only performed for one or two nights. An alternative to assess the quality of sleep is the use of wearables, such as actigraphy devices. Actigraphy devices are wristwatch-like devices that allow continuous activity recording for several weeks. Several studies have tested the efficacy of wrist-worn actigraphy devices for activity and sleep monitoring \citep{weiss10,diaz2015fitbit} and even compared the data with PSG, with satisfying results ~\citep{cole92,jean01b,quante18,smith2018use}.

There have been various studies that have shown how activities carried out during the day could affect sleep. ~\citep{kredlow2015effects} show the impact of different exercise levels on different stages of sleep, sleep time and sleep efficiency. ~\citep{chennaoui2015sleep} discuss how sleep and exercise influence each other: how different physical activity levels influence physical stages during sleep, and how sleep quality influences exercise performance. Other studies demonstrated that a certain amount of exercise helps to reduce insomnia~\citep{yang2012exercise, merrill2007effects} and sleep apnea~\citep{awad2012exercise}. 
%All of these studies conducted their experiment on particular groups of people using direct monitoring, surveys, questionnaires or using PSG.

%An alternative to assess the quality of sleep is the use of wearables, such as actigraphy devices. Actigraphy devices are wristwatch-like devices that allow continuous activity recording for several weeks.

%In the pioneering work of Webster et al.~\cite{webster82}, an early version of actigraphy devices showed the feasibility of using an activity-based sleep monitor system to quantify sleep time with reasonable accuracy while avoiding the impracticality and expense of PSG. While the signals captured by an actigraphy device are not as detailed as the ones captured by a PSG, it still allows the identification of sleep-wake states~\cite{sadeh94}.

With the advent of the Quantified-Self movement \citep{swan2009qs,lupton2016quantified}, which focuses on collecting and analyzing data about oneself, more people have become aware of their health, which in turn has led to a plethora of wearable devices and fitness tracking applications for consumers to improve their health and well-being. The products available are getting more sophisticated and can track a wide array of things, like heart rate, sleep patterns, steps taken, and provide analysis and feedback on how these can be improved to lead a healthier life. These self-tracking devices have empowered people to monitor their health proactively and be able to take appropriate measures to alleviate possible health risks. 

This immense body of work in the domain of wearable devices and sleep tracking motivated us to study the relationship between physical activity and sleep quality using machine learning techniques, and to  find ways of assisting people in improving their sleep. 
%Along with this progress in the health domain, there has been a great deal of progress in the computing domain for efficiently processing huge amounts of data. The evolution of machine learning and deep learning techniques is transforming the field of sleep medicine as new techniques are continually emerging \citep{hossain2018active}. 
Moreover, most of the existing fitness tracking applications provide generic recommendations to consumers for helping them achieve their health goals, usually based on gender and age. The “one-size-fits-all” principle doesn't necessarily fulfill the health objectives of many people \citep{evenson2015systematic}. 

Personalization is critical for providing effective recommendations, i.e. ach-ieving required behavioral change, as the state of health, level of self-control, physical activity types, duration of those activities, and timing would be different for each subject. Recommendations should be such that they are based on users' health goals, but at the same time are easy to adopt. If the goal is to lose weight, but the user has very high BMI, then recommending 10,000 steps a day may not be an achievable recommendation and may even lead to the user abandoning the recommendation system completely. Instead, recommending 5,000 steps a day, and gradually increasing the goal based on adoption would be the right approach.

Google Maps providing directions to a user to go from place A to B is an example of a personalized recommendation that is based on the user's goal (to reach a destination B). If the application gives wrong directions (maybe a road was closed), then the consequence would at worst be that the user gets late in reaching their destination. But in the health domain, the consequences of incorrect recommendations are far more significant and can be immediate or even long-term. For example, if person A has a resting heart rate of 90 and person B has a resting heart rate of 65. Recommending a running activity to A would increase their risk of stroke, whereas the same recommendation would be good for person B. Therefore recommendations should be adapting and evolving, keeping in mind the health of the user.  

\citep{ni2019modeling, loepp2018recommending} present such personalized systems for recommending workout routes based on user preferences and health profiles. %\citep{pilloni2017recommendation} used behavior patterns to predict if a sportsman would stop exercising, and these predictions are used to add a human in the loop - the sports coach.
\citep{alcaraz2019evolutionary} recommend food and exercise bundles to users based on their preferences and health goals. \citep{nosakhare2020toward} used self-reported health data from users to build machine learning models for predicting stress levels of users each day as well as finding days similar to the current day but with lower stress levels so that recommendations can be made about changes in behaviors that would result in lowering stress.

%TODO: add details about user profiles, good sleep recipes
%highlight novel contributions
\textbf{In this paper}, we present PARIS: a goal-directed Personalized Activity Recommender system for Improving Sleep quality, that uses individual health and lifestyle constraints for making activity recommendations. We are addressing the goal of improving sleep quality, but this approach can be expanded to achieve other personal health goals like lowering heart rate and increasing daily step count. The key contributions of this paper are:
\begin{enumerate}
   \item Providing personalized activity recommendations to users at various points in time, using their current activity and calculating activity deficits to achieve a daily goal. 
   \item An approach to generating activity recipes that would lead to a specific daily goal.
   \item Using health metadata, or lifestyle constraints, to select which recommendations would be most appropriate for a given user. 
\end{enumerate}

%with the goal of improving sleep quality, which is achieved by extracting high confidence, actionable insights from wearable data by applying machine learning techniques. 

%Intro from Saksham's UROP poster

%Specifically, activity recipes are generated for high sleep efficiency from heart rate and activity data. These activity recipes would then be supplied to an activity recommendation engine for providing continuous recommendations to users during their daily routines, with the main objective being the improvement of overall quality of sleep. 
%There has been research done on recommending activities based on recipes learned from clustering over activity patterns of the user data. %TODO: citation.
%However in this paper we focus on data about a person’s current health i.e. heart rate data, to group biologically similar people into clusters and then learn their activity patterns to recommend activities. 

%Intro from Abhiraj's UROP poster
%Futhermore, users on a weekly basis have several behavior modes that their bio-functions can be divided into (for example, weekdays and weekends). The intraday heart rate and step count patterns could reflect users' behavior modes. Performing time series clustering on heart rate and step count data, this project also aims at finding an optimal number of cluster centers which would correlate to most evident behavior modes for a specific user. These modes would again be used to provide appropriate activity recommendations to users.

%% file: 2problem_definition.tex
%A generic overview of the problem that we are trying to contribute solve is, 
For a target user $\mathbf{u_t}$, our research aims to provide a set of personalized activity recommendations at different moments of the day $\mathbf{t_m}$ based on their health and wellness factors as well as their daily activity patterns, what we term as behavior modes $\mathbf{BM}$, calibrated against their activity until the time of recommendation $\mathbf{t_h}$, to assist them in meeting their daily sleep goals.

%\subsection{Notations}
Let $\mathbf{X_u}$ be a matrix of minute level actigraphy data for the entire day for each user $\mathbf{u}$ such that
\begin{equation*}
            \mathbf{X_u}=\left(\mathbf{x}_{u, 1}, \ldots, \mathbf{x}_{u, N}\right) \in \mathbb{R}^{N \times 1440}
\end{equation*}
where $\mathbf{x}_{u, i}$ is the actigraphy data for user $\mathbf{u}$ at day $i$, and $N$ is the number of days for which data is available for user $\mathbf{u}$.

The raw actigraphy counts are also categorized into activity level labels based on their values, such that for $\mathbf{x}_{u, i}$ vector, there is a corresponding  $\mathbf{l}_{u, i}$, which contains totals for each activity level label at day $i$. The activity level label can take one of the values in the $\mathbf{activity\_levels}$ enumeration, where \\
$\mathbf{activity\_levels}$ = $\{Light, Moderate, Sedentary, Vigorous\}$ %and the enum is represented by $\mathbf{activity\_levels}$. 

$Sedentary$ activity level is discarded from this enumeration as this category is not used for activity recommendation. Based on this, let $\mathbf{S_{u, t_1:t_2}}$ represent the aggregated minute count for each activity level label in the $\mathbf{activity\_levels}$ enumeration for user $\mathbf{u}$ from time $\mathbf{t_1}$ to $\mathbf{t_2}$ for each day, i.e. 
\begin{equation*}
    \mathbf{S_{u, t_1:t_2}}=\left(\mathbf{s}_{u, 1, t_1:t_2}, \ldots, \mathbf{s}_{u, N, t_1:t_2}\right) \in \mathbb{R}^{N \times 3}    
\end{equation*}

\subsection{Behavior Modes}
Using the actigraphy time series dataset $\mathbf{X_u}$ for each user, compute their behavior modes for the day, i.e. activity patterns that represent a particular day for the user. We introduce further notation to account for the computed behavior mode as follows:

$\mathbf{X^{b}_u}$ is the modified matrix $\mathbf{X_u}$ which contains only actigraphy data for days which belong to the same behavior mode $\mathbf{b}$. This is defined as 
\begin{equation*}
    \mathbf{X^b_u} = \{\mathbf{x}_{u, i} \in \mathbf{X_u} | i \in \mathbf{BM_{u, b}}\}
\end{equation*} 
where $\mathbf{BM}_{u, b}$ is a set of days that belong to the same behavior mode $\mathbf{b}$ for user $\mathbf{u}$. We will use a similar notation for aggregated activity level minute count represented as $\mathbf{S^{b}_{u, t_1:t_2}}$.

\subsection{Activity Recipes}
Using the aggregated activity level label summaries for the whole day $\mathbf{S^{b}_{u}}$ for each behavior mode $\mathbf{BM_{u}}$ for a user $\mathbf{u}$, compute the activity recipes that lead to healthy sleep for the user. 
Activity recipes is a matrix represented as
\begin{equation*}
    \mathbf{A^{b}_{u}}=\left(\mathbf{a}^{b}_{u, 1}, \ldots, \mathbf{a}^{b}_{u, L}\right) \in \mathbb{R}^{L \times 3}
\end{equation*}    
It represents $L$ different activity recipes learned for each behavior mode for each user and comprises the total aggregated minutes for each activity level label in $\mathbf{activity\_levels}$ enumeration.

\subsection{Activity Recommendation}
Let $\mathbf{M_u}$ be a vector $\in \mathbb{R}^{m}$ containing information $\mathbf{m}$ on different statistics like age, gender, BMI, etc., that help profile the user's health and wellness aspects. 

Let $\mathbf{t_m}$ represent the minute count since the onset of day (12:00 AM) when we provide a recommendation. At $\mathbf{t_m}$, a target user $\mathbf{u_t}$'s activity vector is represented as $\mathbf{x_{u_t, 0:t_m}}$
%$\mathbf{S_{u_t, 0:t_m}}$. 

The recommendation problem our research is attempting to solve is to (i) find the behavior mode $\mathbf{b}$ that is closest to target user $\mathbf{u_t}$ and then (ii) find activity recipe $\mathbf{a}^{b}$ that is most similar to the aggregated activity level vector $\mathbf{S_{u_t, 0:t_m}}$ while taking into account lifestyle constraints $\mathbf{M_{u_t}}$.

%Build a personalized activity recommendation engine that will recommend what activities to perform for how much time based on the identified behavior mode for the day for the user, closest activity recipe and user's health and wellness information $\mathbf{M_u}$ at different moments of time in the day to help the user achieve their sleep goals.

%% file: 3methodology.tex
\begin{figure*}
  \includegraphics[width=1\textwidth]{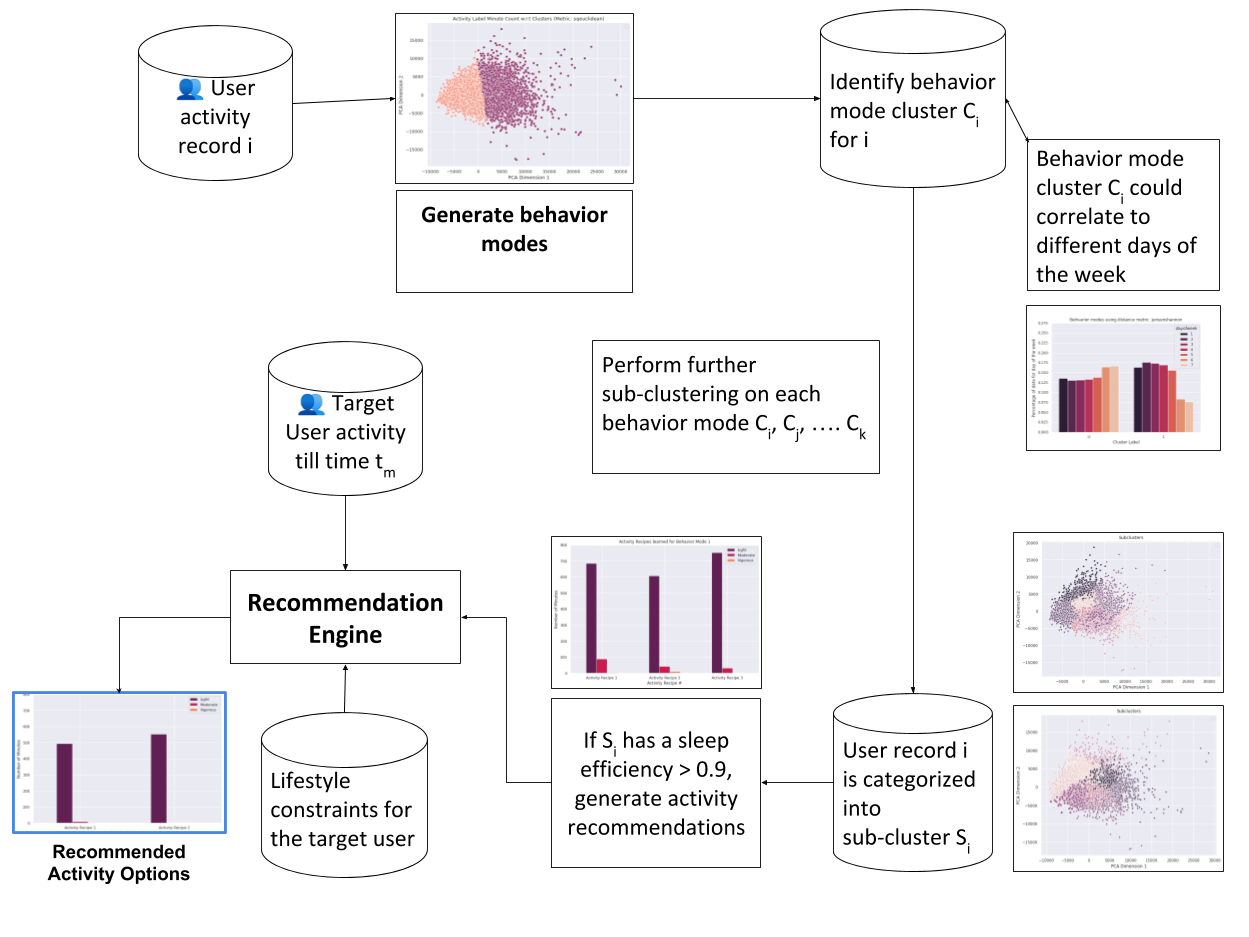}
  \caption{Activity Recommendation System} 
  \label{fig:architecture}
\end{figure*}

The main tasks performed as part of our research and listed below, and the process flow is shown in Fig. \ref{fig:architecture}.

%included 
%building a data collection system to get minute level data from wearable devices, 
%creating profile clusters and generating behavior modes for each subject based on their %heart rate or 
%activity throughout the day using step count, developing predictive models to extract activity recipes \citep{sathyanarayana2017} for each behavior mode within each profile cluster, and finally using these models to recommend at any point in time, the activities the subject should carry out for the rest of the day, to ensure a good night's sleep. The steps 

\begin{enumerate}
    % \item Building a data collection system to get minute level data from wearable devices, which include steps, calories, activity levels, heart rate and sleep %and preprocessing it for analysis
    \item Generating behavior modes using clustering algorithms for each subject based on their activity (using movement intensity) (Section \ref{sec:behaviormode})
    \item Developing predictive models to extract various activity recipes \citep{sathyanarayana2017} for each behavior mode, which result in \textit{Good Sleep} (Section \ref{sec:activityrecipe}.%for each individual behavior mode identified
    \item Using these models to recommend at any point in time, the activities the subject should carry out for the rest of the day, to ensure a good night's sleep (Section \ref{sec:activityrec})
    %\item Updating the recommendation engine based on user compliance to activity recommendation and actual sleep metrics collected
\end{enumerate}

%, which was used, along with age, gender and BMI, to group people into profile clusters using using the K-Means clustering algorithm.

\subsection{Generating Behavior Modes} \label{sec:behaviormode}
%TODO replace figures with those for hchs data

For all subjects, we performed clustering over the minute-level activity count time series. %and heart rate time series data. 
Each cluster represents the subject's routine based on activity performed throughout the days for the same subject and we label these clusters as the subject's behavior modes. The cluster centroids explain the aggregate characteristics of activity time series in their respective clusters thus describing the behavior mode.
%TODO: add reference to the figure

Clustering performed for detecting behavior modes used K-Means clustering, although we did use many different distance metrics to compute similarity/distance between time series for our clustering:

\begin{enumerate}
    \item L1 norm(X,Y) \begin{math} = \sum_i |x_i - y_i| \end{math}
    \item L2 norm(X,Y) \begin{math} = \sqrt{\sum_i |x_i - y_i|^2} \end{math}
    \item Dynamic Time Warping (DTW)
    \item Correlation(X,Y) \begin{math} = \dfrac{\sum X Y - n \bar{X}\bar{Y}}{\sqrt{\sum X^2 - n \bar{X}^2}\sqrt{\sum Y^2 - n \bar{Y}^2}} \end{math}
    \item Kullback-Leibler (KL) Divergence(X,Y)  \begin{math} = \sum_i x_i \times \log(\dfrac{x_i}{y_i}) \end{math}
    \item Jensen-Shannon (JS) Divergence(X,Y)
    \begin{math} = \frac{1}{2} KL(X,M) + \frac{1}{2} KL (M,Y)\end{math} where 
    M = \begin{math} \frac{1}{2}(X+Y)\end{math}
\end{enumerate}

We considered a variety of different distance metrics because the concept of distance in higher dimensions, especially time series, becomes more abstract and it is important to have a stable yet powerful. While L1 Norm and L2 Norm are on the simple side of the spectrum, metrics like DTW and KL divergence are much more powerful in extracting information based on overall time series distribution. KL and JS are very similar distance metrics. JS can be considered a more symmetric version of KL since it is not symmetric. We computed KL divergence X$\to$Y and Y$\to$X and took the mean of the two. JS divergence takes the mean of X and Y and takes the mean of KL divergence between X and its mean and Y and its mean.

%\textcolor{red}{TODO: silhouette score formula and explanation}

We used silhouette score as the primary evaluation criteria for our clustering results for each clustering model from our grid search. We optimized for the maximum silhouette score to find the best clustering model. We used silhouette score because it helps maximize intra-cluster similarity and minimize inter-cluster similarity, thus helping group similar activity pattern while separating dissimilar patterns. The silhouette score for a single point is described as follows

\begin{displaymath}
    S(i) = \frac{b(i) - a(i)}{max[b(i), a(i)]}
\end{displaymath}

Here, a(i) is defined as the mean distance between the data point i and every other data point in the same cluster. And b(i) is defined as the mean distance between data point i and all data points from its closest neighbor cluster. 

%was this done for HCHS data? It was but we did not really get any different results. 
Apart from clustering over minute level time series data we also perform similar clustering over principal frequencies identified for each time series using FFT to reduce the noise associated. Clustering over the frequency domain makes sure that we group behavior mode based on core characteristics. From the behavior modes that we obtain, we can find the weekday distribution to better understand the weekly routine and the weekly behavior mode pattern.

%Saksham
%TODO: check and add/modify details if needed
\subsection{Activity Recipes for Good Sleep}\label{sec:activityrecipe}
After computing the behavior modes for each subject, activity recipes were learned for \textit{Good Sleep} per behavior mode per cluster. Clustering was applied again over aggregated activity level summaries for each day within the same behavior mode. Activity level summaries were calculated as the duration (number of minutes) of a specific activity level label (Light, Moderate and Vigorous) from the time the subject wakes up to the current time. For this step, we removed the sedentary activity level because our recommendation engine would recommend being physically active, which only depends on light, moderate and vigorous activity levels. For each record within the same cluster, we computed the sleep efficiency which is defined as the ratio of total time asleep to total time in bed. 

\begin{displaymath}
    Sleep\ Efficiency = \frac{Total Minutes Asleep}{Total Minutes In Bed} = 1 - \frac{Minutes Awake In Bed}{Total Minutes In Bed}
\end{displaymath}

Then each record was tagged as either \textit{Good Sleep} day or \textit{Poor Sleep} day based on the sleep efficiency such that \textit{Good Sleep} is defined as sleep efficiency greater than 0.90. This process was repeated for each cluster identified for each behavior mode for each subject. Each resulting cluster was identified as a \textit{Good Sleep} cluster if the ratio of the number of records with \textit{Good Sleep} within the cluster to the number of records with \textit{Poor Sleep} within the cluster is greater than or equal to 2. After identifying all the \textit{Good Sleep} clusters, activity recipes were calculated as the cluster centers for those sub-clusters.

For detecting behavior modes the clustering algorithm selected was K-Means clustering using L2-Norm distance metric, such that we optimized the silhouette score to find the number of sub-clusters. The data for this step consisted of independent features in a 3-dimensional space (activity levels excluding sedentary activity) which is a perfect candidate for K-Means clustering because of the low number of dimensions and feature independence. Since the features are in an N-Dimensional space and do not represent any kind of time-series or distribution, we selected L-2 norm as the distance metric.

%TODO cite Aarti's thesis?

\subsection{Continuous Activity Recommendation Engine}\label{sec:activityrec}
%\textcolor{red}{TODO: Meghna - Reword this section. Looks confusing}
The activity recommendation process assigns a behavior mode based on the target user $\mathbf{u_t}$'s activity till time $\mathbf{t_m}$, i.e., $\mathbf{x_{u_t, 0:t_m}}$. To find the appropriate behavior mode, $\mathbf{x_{u_t, 0:t_m}}$ is compared with the cluster centers of the generated behavior modes, and the cluster center with the shortest distance is selected as the behavior mode $\mathbf{b}$. Comparison is done on cropped data to resemble the amount of data collected so far, i.e. for time duration $(0,{t_m})$. 

Next, $\mathbf{x_{u_t, 0:t_m}}$ is converted to the activity level vector $\mathbf{l_{u_t, 0:t_m}}$, which is then aggregated by activity levels to get $\mathbf{S_{u_t, 0:t_m}}$. $\mathbf{S_{u_t, 0:t_m}}$ is then compared to the cluster centers of the activity sub-clusters, which returns the probability of  $\mathbf{S_{u_t, 0:t_m}}$ belonging to each cluster. The \textit{Good Sleep} activity recipes $\mathbf{a_b}$ corresponding to the cluster centers are ordered by this probability, and are used to calculate activity deficit for the remainder of the day. 
%in order to ensure the sleep that followed would be categorized as good sleep.  

%Comment from Meghna: which data pipeline are you talking about?
%Activity recommendation is based on running the activity count time series data through the original predictive model pipeline such that we find the behavior mode most aligned to the current day, based on the proximity to the cluster centers.% of behavior modes identified for the particular user. Comparison is done on cropped data to resemble the amount of data collected so far. 
%Once the behavior mode is identified, we calculate the aggregated activity level minute counts until the current time and find the probability (soft labels) of the current day belonging to the good sleep clusters %originally identified during training. The activity summaries are compared to the cluster centers for a particular time so that there is no discrepancy. 
The model then recommends different activity regimes based on the deficit to achieve the same activity counts as being in the activity recipes.%in order of decreasing probabilities for each soft cluster label. 
The model also takes into consideration the user's metadata $\mathbf{M_{u_t}}$ to reorder the activity recipes if required, to ensure that a recommendation is such that it does not have adverse effects on the user's health, along with being one that could be easily adopted. %which are greater than some tuned probability threshold.

% pipeline over the newly collected user data from the Fitbit. Recommendation is made at different moments of the day using the data collected so far. The process starts by finding the closest behavior mode using distance metric used for the clustering. After learning the behavior mode we find the closest activity summary feature within this behavior mode using the same distance metric which was used for sleep recipe collection. Once it is determined which sleep group the data collected so far resembles, we recommend based on the difference of the activity level summary learned and the current activity level summary.

%Meghna: I have rewritten the steps here - check and verify that I am not missing anything important.

% The activity recommendation process found the closest behavior mode based on current day's activity thus far. Next, the activity summary feature that most resembled current activity was selected, which in turn was used to recommend subsequent activities for the remainder of the day in order to ensure the  sleep that followed would be categorized as good sleep.  

%% file: 4expts_results.tex
\subsection{HCHS/SOL Dataset}
Dataset collected for the Hispanic Community Health Study / Study of Latinos (HCHS/SOL), and made available by NSRR, has been used in this project \citep{zhang18, redline2014sleep}. 
%The dataset used for this study is the Hispanic Community Health Study / Study of Latinos (HCHS/SOL) dataset \cite{zhang18, redline2014sleep} available from sleepdata.org. This dataset is a population-based epidemiological study of Hispanic/Latino populations. %to determine the role of acculturation in the prevalence and development of disease. 
The dataset contains data for Latino adults aged 18–74 years at enrollment, which includes one night of in-home PSG data for 12,088 participants, along with aggregated survey questionnaire data about health, lifestyle, and sleep.
%Heart rate time series is extracted from this dataset. %Participants self-identified as Hispanic or Latino and of Mexican, Cuban, Puerto Rican, Dominican, Central/South American, or other or more than one Latino heritage. A baseline clinical examination and interview were conducted and included demographic and biological assessments (e.g., anthropometrics, blood draw).
Additionally, the Sueño Ancillary study recruited 2,252 HCHS/SOL participants to wear wrist-worn actigraphy devices (Actiwatch Spectrum, Philips Respironics) for a week. Actigraphy data for 1,887 participants has been made available for use.
The study was approved by the Institutional Review Boards at all HCHS/SOL institutions and written informed consent was obtained from all participants.

Our research focuses on utilizing the metadata about each subject along with heart rate and actigraphy time-series data to be able to detect behavior modes and extract sleep recipes. The heart rate data for subjects are from the one-night PSG data. The actigraphy data includes about 7 days of activity data (activity count) at 30-second intervals along with annotations for sleep and wake periods for 1,887 subjects. The intersection between these two cohorts was a dataset of 1,782 subjects. Metadata for each subject includes many biological, biographical, and physical features like age, gender, BMI, diabetes, ECG abnormalities, medical history, etc. 

The actigraphy data has been annotated by the Actiwatch software along with data curators, such that each row has an interval type, which can be ACTIVE, REST, REST-S or EXCLUDED. A subject is considered asleep between the REST intervals where the REST records are before sleep onset and after sleep offset while REST-S is the actual sleep period. There is also an awake indicator field, which is 1 when the subject is awake.

%\textcolor{red}{TODO:Meghna  features selected for use at the time of recommendation}

\subsection{Building Feature Space}
Activity data of 1,782 subjects were split into 24-hour periods and then data points aligned by their time. The data is aggregated to minute-level counts using a rolling sum with a window of size 2. Any subject with less than 7 days of activity was dropped. The resulting dataset consisted of 1,769 subjects with 7 activity time series each, where each time series corresponded to one day of the week. This data will be used to generate behavior modes (section \ref{sec:behaviormode})
%12383 rows

%\begin{figure*}
 % \includegraphics[width=0.85\textwidth]{img/centers_euclidean.png}
%\caption{Cluster centeroids for behavior modes using Euclidean} 
%\label{fig:recipes}       
%\end{figure*}

The minute-level counts were next used to generate activity labels based on activity level thresholds or cut-points \citep{colley2011physical, wong2011actical}, where the labels correspond to movement intensity levels: sedentary, light, moderate or vigorous. Only the rows which had been labeled ACTIVE were selected for this step. The counts were aggregated per subject per day per activity label to generate activity level summaries. A sample resulting data is shown in table \ref{tab:activitylevelsample}. This data will be used for generating activity recipes (section \ref{sec:activityrecipe})

	\begin{table}[h] 
	\renewcommand{\arraystretch}{1.5}
	%\centering 
		\caption{Sample activity level summaries}
		\label{tab:activitylevelsample}
		\begin{tabular}{ r  r  l r}			
        \hline\noalign{\smallskip}
		{\textbf{Subject Id}} & 
		{\textbf{Day of the week}} & 
		{\textbf{Activity Label}} & 
		{\textbf{Minutes}}  \\ 
    	\hline\noalign{\smallskip}
		1 &  1 & light & 880 \\ 
		1 &  1 & moderate & 4 \\ 
		1 &  1 & sedentary & 556 \\ 
		1 &  2 & light & 828 \\ 
		1 &  2 & moderate & 5 \\ 
		1 &  2 & sedentary & 607 \\ 
		\hline\noalign{\smallskip}
		\end{tabular}
    \end{table} 

%\textcolor{red}{TODO:Meghna - add how sleep part of the data was selected, sleep efficiency calculation}
%TODO: Add WASO (wake after sleep onset) to the formula 
For calculating sleep duration and sleep efficiency, only the rows labeled REST and REST-S were used. The awake indicator was used to check if there are Wake After Sleep Onset (WASO) periods if the awake indicator was 1 for at least 5 minutes (i.e., ten consecutive rows of data). The total minutes of WASO and REST are considered as $MinutesAwakeInBed$ during the sleep period.

%Metadata
%30 features were selected and the values for each subject saved.

%The data was pre-processed by removing null values using means found in hourly segments corresponding to each subject for all time series. 
% Next, the time series data was smoothed using a seasonal decomposition using moving averages. A moving average window of 2 hours was used for heart rate and of 15 minutes for steps.

\begin{figure*}
  \includegraphics[width=1\textwidth]{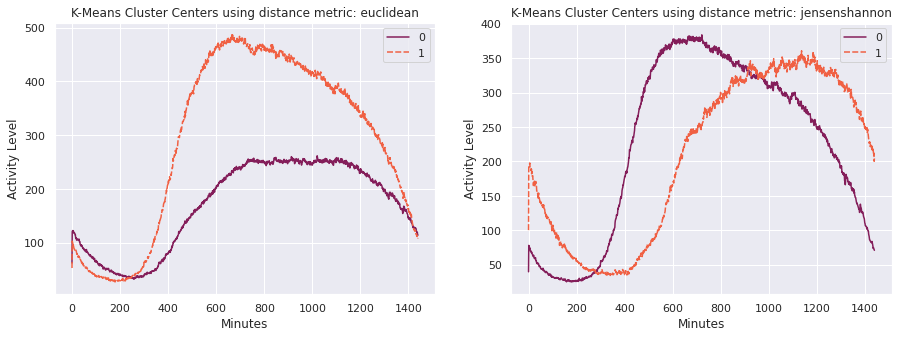}
\caption{Cluster centroids for behavior modes using distance metrics Euclidean and Jensen-Shannon (JS)} 
\label{fig:cluster_centers}       
\end{figure*}

The activity level data was also transformed using Fast Fourier Transform (FFT) to get the principal frequencies and migrate the data from time to frequency domain. We used FFT to reduce the time series into a sum of sinusoidal components. Doing so allowed us to reduce the noise associated with a time series. We took the first 25 components to further eliminate noise and outlier values from the time series.

%Abhiraj
\subsection{Generating Behavior Modes}
\begin{figure*}
  \includegraphics[width=1\textwidth]{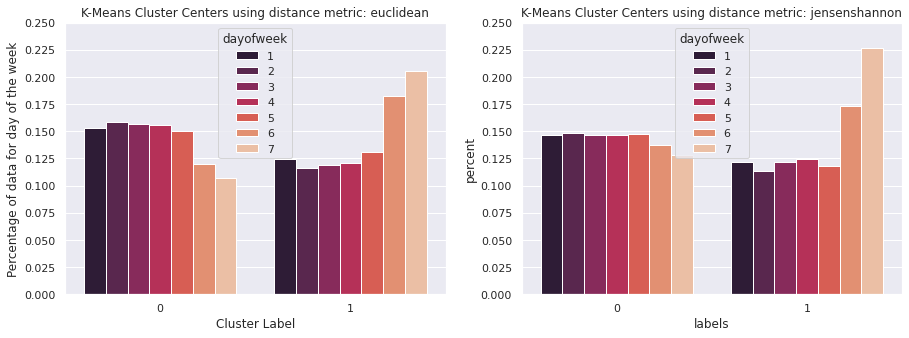}
  \caption{Cluster Composition for behavior modes using distance metrics Euclidean and Jensen-Shannon (JS)} 
  \label{fig:hr_stepcount}
\end{figure*}

%TODO update figure from hchs behavior modes
For user behavior modes, the number of clusters with the highest silhouette score was found to be two. We ran twenty scenarios to compare silhouette scores and the highest we achieved was with cluster centers being equal to two. 
The purity of all such clusters was tested to find the composition based on the days of the week.
Fig. \ref{fig:hr_stepcount} shows the activity cluster composition using two distance metrics: Euclidean and Jensen-Shannon (JS) divergence.  Clustering over activity using Jensen-Shannon led to a marginally higher cluster purity compared to clustering done with Euclidean. Cluster 1 in Jensen-Shannon clustering has a higher number of weekend days compared to cluster 1 in euclidean clustering. We used Jensen-Shannon over Kullback-Leibler because Jensen-Shannon is a symmetrical incarnation of the same formula.
Figure \ref{fig:cluster_centers} describes the cluster centers for the two behavior mode clusters. Each center gives us insight into what user behavior looks like on weekdays or weekends. 
 Figure \ref{fig:hr_stepcount} shows the two cluster composition.

\subsection{Activity Recipes for Good Sleep}
\begin{figure}
  \includegraphics[width=1\textwidth]{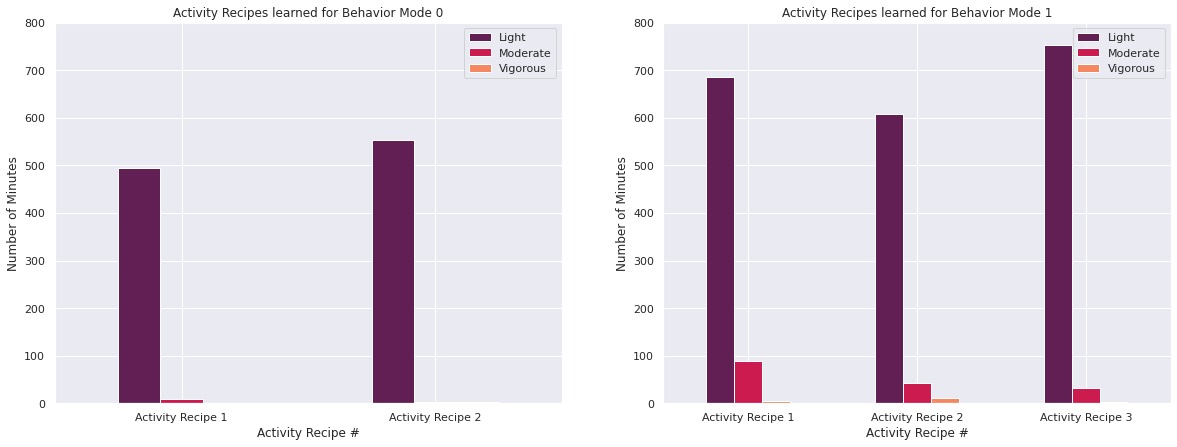}
\caption{Activity recipes for good sleep generated for a specific behavior mode} 
\label{fig:recipes}       
\end{figure}

Based on the computed behavior modes, when we ran sub-clustering to extract activity recipes for each user we found a varying number of activity recipes across the users. Fig. \ref{fig:recipes} shows the different activity recipes generated for each behavior mode,  with varying levels of light, moderate and vigorous activities. These figures accurately depict that activity recipes have a strong correlation with the behavior modes. That is activity recipes for behavior modes with less overall activity also have fewer minutes for moderate and vigorous activity. %while behavior modes which show more activity throughout the day.
%\textcolor{red}{TODO: add an example of behavior and sleep recipe to make sure reader understands what exactly is good sleep recipe}
%TODO Insert sleep recipes here

% \begin{table}
% 		\caption{Comparison of Distance Metrics over Cluster Purity}
% 		\label{tab:accuracy}
% 		\begin{tabular}{ l  l p{3.4cm}}		
% 		\hline\noalign{\smallskip}
% 		{\textbf{Distance Metric}} & 
% 		{\textbf{Cluster Purity}} & 
% 		{\textbf{Avg. Activity Recipe Cluster Purity}}    \\ 
%         \noalign{\smallskip}\hline\noalign{\smallskip}

% 		L1 Norm &  0.5473 & 2.13 \\ 
% 		L2 Norm &  0.5848 & 2.34 \\ 
% 		DTW & 0.6372 & 2.59  \\ 
% 		Correlation &  0.6516 & 3.12 \\ 
% 		KL Divergence & \textbf{0.7006} & 4.18 \\ 
% 		\noalign{\smallskip}\hline
% 		\end{tabular}
% \end{table} 

%\begin{figure*}
%  \includegraphics[width=\textwidth]{img/KLdivergence.png}
%\caption{Clustering Heart Rate and calories using KL divergence} 
%\label{fig:kldivergence}       
%\end{figure*}

%TODO: Details of the other clustering techniques

To determine activity recipes, various distance metrics were used for clustering and 
Euclidean produced the best results among all the clustering techniques, i.e. 
the purest clusters for good and poor sleep. Based on these clusters, good sleep recipes were determined, which indicate the different activities and the duration of each of those that should be carried out to ensure good sleep. %Figure \ref{fig:recipes} shows the generated sleep recipes with varying levels of light, moderate and vigorous activities.

\begin{figure}
  \includegraphics[width=1\textwidth]{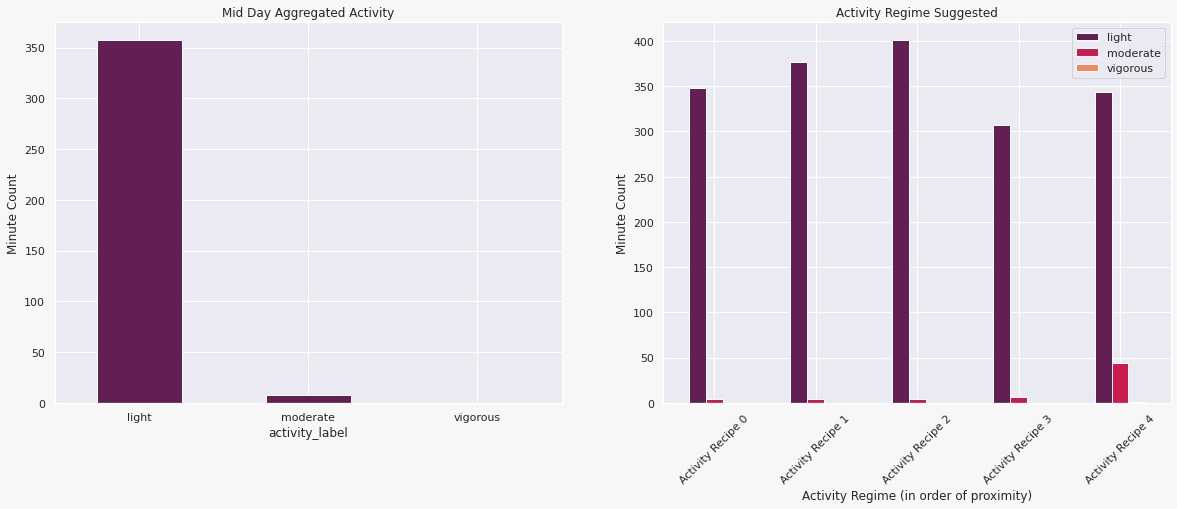}
\caption{Activities recommended based to a target user based on their activity till time $t_m$} 
\label{fig:activityrec}       
\end{figure}

%TODO results for recommender
Since this work was not on streaming data, online evaluation could not be performed. As it was an offline evaluation, we could not test how the subjects would perform in real-time, i.e. whether they would follow the given recommendation or not, and what the outcome was if the recommendation was followed, versus when it was not followed. 
Instead, retrospective evaluation \citep{sathyanarayana2017} was used to find subjects with the closest activity pattern and the quality of subsequent sleep to decide if the recommendation was successful or not.  Fig. \ref{fig:activityrec} shows one such set of recommendations that were generated based on the target user's activity time series till the middle of the day. The left figure is for the target user's existing activity categorized into activity labels, while the right figure presents a list of activity recommendations, ordered by their proximity to the input data.

%% file: 5discussion.tex
%Discussion
%\subsection{Impact}
%\textcolor{red}{TODO (Meghna): expand this section}
%This work will provide a system to be used by individuals for themselves, and care providers (doctors, nurses, family) for their subjects for improving their physical and mental health through personalized recommendations for better sleep.

\subsection{Initial Hypothesis}
Our team has been working on an activity recommendation engine for quite some time and has progressed slowly from a purely hypothetical design to data-driven development. Our initial hypothesis for building the activity recommendation engine was to group similar people (clustering) based on their cardiovascular health using their Heart Rate time series data for the day and then sub clustering these groups based on similar aggregated activity levels to learn activity recipes for good sleep. This hypothesis was then extended to include a parallel pipeline that learns the behavior modes for each person based on their activity time series for the whole day and performs similar sub clustering. The activity recommendation engine then evolved to recommend activities based on an overall evaluation of sleep recipes determined for each pipeline using their distances from cluster centers. With the updated model, we could provide personalized activity recommendations while making sure to fall back to generalized activity recommendations based on other users if there was an anomalous day for that user.

\subsection{Alternative Data Source}
To confirm the validity of our hypothesis, we built the activity recipe extractor model over our custom Fitbit data. We built a data hosting server that periodically collected minute-level heart and activity data from Fitbit watches worn by students who volunteered to participate in our research. Once we collected enough data (around 4 months' worth), we were successfully able to identify behavior modes for all the users. It was verified by the subjects that behavior modes identified coincided with their weekly schedules. Apart from this, heart rate clustering also helped successfully group similarly aged people into clusters. We also learned various activity recipes which seemed to be in line with the students' schedules over the semester. This experiment helped reaffirm our hypothesis.%boost our confidence in our hypothesis. 
However, the dataset that we worked with was not large enough to build a robust model on.%: only 4 people over 4 months. 
Hence we decided to explore other datasets that have actigraphy data along with heart rate data available for a sufficiently large group of people to build our activity recommendation engine. We then continued our efforts using the HCHS/SOL dataset. %provides us with all the relevant information needed to build this activity recommendation engine. That is when we decided to continue our efforts on the HCHS dataset.

\subsection{Heart Rate Time Series Data}\label{sec:hrclustering}
After moving to the HCHS/SOL dataset, we found out was that the heart rate PSG data for the users corresponded to their baseline visit and was for a single sleep session. The baseline visit for the users also did not align with their actigraphy data. Due to the nature of this dataset, we removed the first step in our original activity recommendation pipeline - clustering users based on similar cardiovascular health.
%we removed the generic activity recommendation pipeline of the mode which was based on clustering users based on similar cardiovascular health.

\begin{figure*}
  \includegraphics[width=0.85\textwidth]{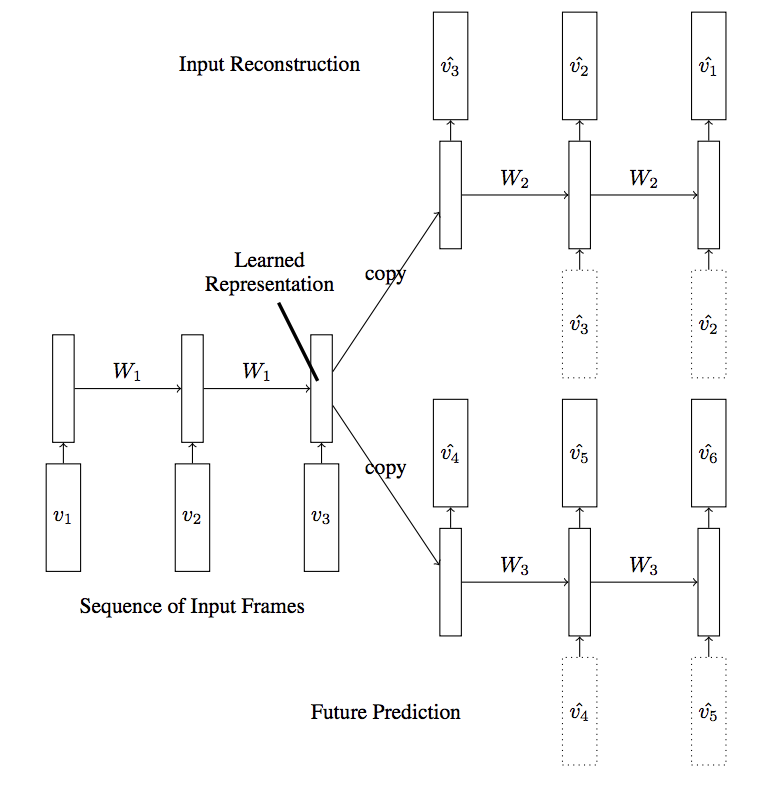}
\caption{The Composite LSTM Autoencoder Model: it predicts the future as well as the input sequence.} 
\label{fig:composite_lstm_autoencoder_arch}       
\end{figure*}

\begin{figure*}
  \includegraphics[width=0.85\textwidth]{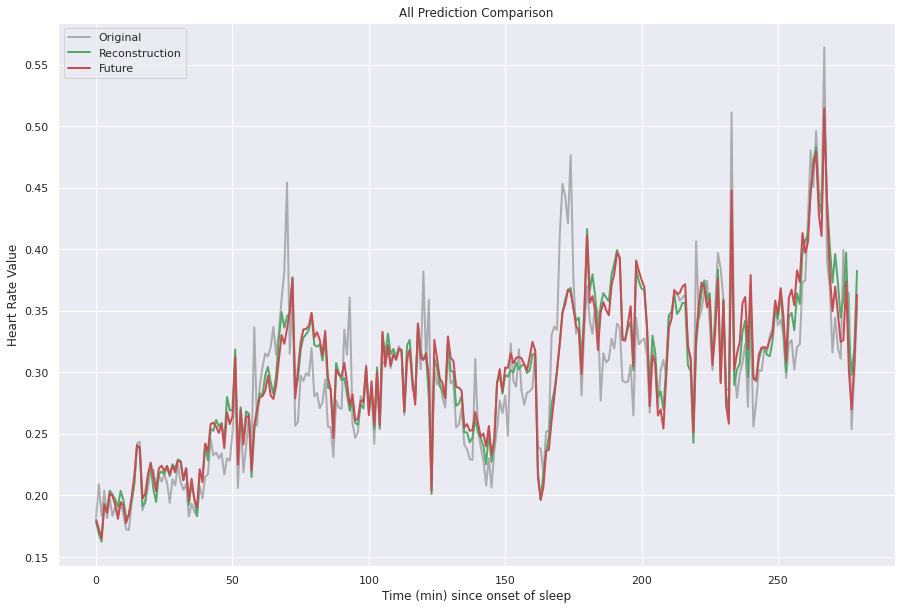}
\caption{Prediction made by our composite LSTM Autoencoder model.} 
\label{fig:composite_lstm_autoencoder_prediction}       
\end{figure*}

\subsection{Deep Learning}
In our project, we evaluated the performance of many different clustering algorithms using different distance metrics over the simple time-series data (minute level), and over the principal frequencies of the time series obtained using FFT. However, we also performed some feature extraction from the time series using an LSTM Autoencoder. LSTM Autoencoder helped us reducer the dimensionality of the minute-level time series and extract meaningful information. This was done because clustering in higher dimensions is much more complex. Using LSTM Autoencoders also helps us learn features independent of each other, unlike time series, thus making sure simple distance metrics like L2 norm are effective in finding similarity and dissimilarity among the records.

The technique we used was a Composite LSTM Autoencoder \cite{composite_lstm_autoencoder} which uses a repeated LSTM unit to reconstruct the entire sequence as well as predict the future sequence. Architecture for the composite LSTM model can be seen in Fig. \ref{fig:composite_lstm_autoencoder_arch}. Training for the model was done to minimize the overall mean squared error (MSE) loss for these parts (reconstruction and prediction units). 

After training our LSTM model over a 60:20:20 split of the dataset as training, validation, and test for 300 epochs, the MSE error over the test set was down to 0.005. Example of a prediction for reconstruction and future values can be found in Fig.  \ref{fig:composite_lstm_autoencoder_prediction}. One thing to note is that the heart rate values were scaled using a MinMaxScaler to bound them to a range of [0, 1] to avoid the problem of vanishing/exploding gradient.

However, the clustering results for the extracted features resembled the results that we obtained from clustering over the raw time-series data. We think that because the number of records for our training was quite low compared to what a model like this expects, it had overfitted the data due to which the learned features resulted in clustering similar to the non-deep methods for clustering (section \ref{sec:hrclustering}).

%% file: 6conclusion.tex
In this paper, we presented a novel approach for goal-directed personalized activity recommendations that finds activities most closely aligned with a user based on activities done thus far, as well as any health constraints the user might have. 

This work will provide a system to be used by individuals for themselves, and care providers (doctors, nurses, family) for their subjects for improving their physical and mental health through personalized recommendations for goals tailored to the end-user.

Based on experiments on the HCHS/SOL dataset, we concluded that subjects had two behavior modes based on the number of clusters generated, which reinforced our initial hypothesis that people generally have a cyclic behavior that varies on weekdays and weekends \citep{pierson2018modeling}.%Heart rate clustering yielded higher cluster purity when KL Divergence is used, while steps clustering yielded better within-cluster purity when clustering is done using FFT transformations. We also find that clustering over heart rate gives us more distinguishable results. This could be due to the nature of step count data and how unreliable it is. The next steps would be to compare distance measures with DTW and use purity-based cluster selection methods.

%Saksham
%From this project we can conclude that using KL Divergence proved to be the best distance metric for clustering heart rate trends to separate good sleep days from bad sleep days. Also, we find that
Using a clustering-based approach helped us in finding numerous activity recipes for good sleep, which denoted different types of schedules. The activity recipes included a mix of intense and light activities that led us to conclude that people could have a variety of activity types, and varying amounts of these activities could all lead to good sleep.% intense activity days to relaxed days. This indicates that a different day can lead to a different type of activity recipe which would lead to good sleep which is expected. Also using heart rate as the principal feature vector for clustering we are able to find clusters that contained very different activity recipes together hence enriching the recipes learned.

%Good sleep has similar daily activities, while poor sleep can vary
%Clustering over heart rate leads to clusters containing varied activity levels helping learn myriad different activities from the same cluster
%Among the distance metrics, KL divergence performs best in finding the purest cluster with respect to sleep efficiency and produces meaningful and varied activity recipes. 

%TODO conclusion about rec sys
\subsection{Future Work}
%Meghna: expand this section
Ongoing work includes adding a feedback loop to our recommendation engine that will continuously recommend activities to users (for example, every 2 hours) and using actions of the users upon receiving the recommendation for subsequent recommendations.

%validation of sleep recipes and the correlation of activity intensity levels to sleep stages (deep sleep, REM sleep, etc.) the subsequent night.
Retrospective evaluation will also be used on existing data to test whether the activity recipe recommendations resulted in good sleep. Another direction for this research is running experiments with other goals.

We also want to explore different clustering algorithms in the future which are better suited for handling high dimensional datasets just like the time series dataset we have. Finally, we want to invest some time in other ways of feature extraction to complement the clustering.

%% file: main.bbl
\begin{thebibliography}{32}
\providecommand{\natexlab}[1]{#1}
\providecommand{\url}[1]{{#1}}
\providecommand{\urlprefix}{URL }
\expandafter\ifx\csname urlstyle\endcsname\relax
  \providecommand{\doi}[1]{DOI~\discretionary{}{}{}#1}\else
  \providecommand{\doi}{DOI~\discretionary{}{}{}\begingroup
  \urlstyle{rm}\Url}\fi
\providecommand{\eprint}[2][]{\url{#2}}

\bibitem[{Alcaraz-Herrera and Palomares(2019)}]{alcaraz2019evolutionary}
Alcaraz-Herrera H, Palomares I (2019) Evolutionary approach for’healthy
  bundle’wellbeing recommendations. In: HealthRecSys@ RecSys, pp 18--23

\bibitem[{Awad et~al(2012)Awad, Malhotra, Barnet, Quan, and
  Peppard}]{awad2012exercise}
Awad KM, Malhotra A, Barnet JH, Quan SF, Peppard PE (2012) Exercise is
  associated with a reduced incidence of sleep-disordered breathing. The
  American journal of medicine 125(5):485--490

\bibitem[{Blume et~al(2020)Blume, Schmidt, and Cajochen}]{blume2020effects}
Blume C, Schmidt MH, Cajochen C (2020) Effects of the covid-19 lockdown on
  human sleep and rest-activity rhythms. Current Biology

\bibitem[{Chennaoui et~al(2015)Chennaoui, Arnal, Sauvet, and
  L{\'e}ger}]{chennaoui2015sleep}
Chennaoui M, Arnal PJ, Sauvet F, L{\'e}ger D (2015) Sleep and exercise: a
  reciprocal issue? Sleep medicine reviews 20:59--72

\bibitem[{Cole et~al(1992)Cole, Kripke, Gruen, Mullaney, and Gillin}]{cole92}
Cole RJ, Kripke DF, Gruen W, Mullaney DJ, Gillin JC (1992) Automatic sleep/wake
  identification from wrist activity. Sleep 15(5):461--469

\bibitem[{Colley et~al(2011)Colley, Garriguet, Janssen, Craig, Clarke, and
  Tremblay}]{colley2011physical}
Colley RC, Garriguet D, Janssen I, Craig CL, Clarke J, Tremblay MS (2011)
  Physical activity of canadian adults: accelerometer results from the 2007 to
  2009 canadian health measures survey. Health reports 22(1):7

\bibitem[{Diaz et~al(2015)Diaz, Krupka, Chang, Peacock, Ma, Goldsmith,
  Schwartz, and Davidson}]{diaz2015fitbit}
Diaz KM, Krupka DJ, Chang MJ, Peacock J, Ma Y, Goldsmith J, Schwartz JE,
  Davidson KW (2015) Fitbit{\textregistered}: An accurate and reliable device
  for wireless physical activity tracking. International journal of cardiology
  185:138--140

\bibitem[{Evenson et~al(2015)Evenson, Goto, and
  Furberg}]{evenson2015systematic}
Evenson KR, Goto MM, Furberg RD (2015) Systematic review of the validity and
  reliability of consumer-wearable activity trackers. International Journal of
  Behavioral Nutrition and Physical Activity 12(1):159

\bibitem[{Jean-Louis et~al(2001)Jean-Louis, Kripke, Mason, Elliott, and
  Youngstedt}]{jean01b}
Jean-Louis G, Kripke DF, Mason WJ, Elliott JA, Youngstedt SD (2001) Sleep
  estimation from wrist movement quantified by different actigraphic
  modalities. Journal of neuroscience methods 105(2):185--191

\bibitem[{Kasasbeh et~al(2006)Kasasbeh, Chi, and
  Krishnaswamy}]{kasasbeh2006inflammatory}
Kasasbeh E, Chi DS, Krishnaswamy G (2006) Inflammatory aspects of sleep apnea
  and their cardiovascular consequences. SOUTHERN MEDICAL JOURNAL-BIRMINGHAM
  ALABAMA- 99(1):58

\bibitem[{Knutson et~al(2006)Knutson, Ryden, Mander, and
  Van~Cauter}]{knutson2006role}
Knutson KL, Ryden AM, Mander BA, Van~Cauter E (2006) Role of sleep duration and
  quality in the risk and severity of type 2 diabetes mellitus. Archives of
  internal medicine 166(16):1768--1774

\bibitem[{Kredlow et~al(2015)Kredlow, Capozzoli, Hearon, Calkins, and
  Otto}]{kredlow2015effects}
Kredlow MA, Capozzoli MC, Hearon BA, Calkins AW, Otto MW (2015) The effects of
  physical activity on sleep: a meta-analytic review. Journal of behavioral
  medicine 38(3):427--449

\bibitem[{Loepp and Ziegler(2018)}]{loepp2018recommending}
Loepp B, Ziegler J (2018) Recommending running routes: framework and
  demonstrator. In: Workshop on Recommendation in Complex Scenarios

\bibitem[{Lupton(2016)}]{lupton2016quantified}
Lupton D (2016) The quantified self. John Wiley \& Sons

\bibitem[{Merrill et~al(2007)Merrill, Aldana, Greenlaw, Diehl, and
  Salberg}]{merrill2007effects}
Merrill R, Aldana S, Greenlaw R, Diehl H, Salberg A (2007) The effects of an
  intensive lifestyle modification program on sleep and stress disorders. The
  journal of nutrition, health \& aging 11(3):242

\bibitem[{Murphy and Peterson(2015)}]{murphy2015sleep}
Murphy MJ, Peterson MJ (2015) Sleep disturbances in depression. Sleep medicine
  clinics 10(1):17--23

\bibitem[{Ni et~al(2019)Ni, Muhlstein, and McAuley}]{ni2019modeling}
Ni J, Muhlstein L, McAuley J (2019) Modeling heart rate and activity data for
  personalized fitness recommendation. In: The World Wide Web Conference, ACM,
  pp 1343--1353

\bibitem[{Nosakhare and Picard(2020)}]{nosakhare2020toward}
Nosakhare E, Picard R (2020) Toward assessing and recommending combinations of
  behaviors for improving health and well-being. ACM Transactions on Computing
  for Healthcare 1(1):1--29

\bibitem[{Palotti et~al(2019)Palotti, Mall, Aupetit, Rueschman, Singh,
  Sathyanarayana, Taheri, and Fernandez-Luque}]{palotti2019benchmark}
Palotti J, Mall R, Aupetit M, Rueschman M, Singh M, Sathyanarayana A, Taheri S,
  Fernandez-Luque L (2019) Benchmark on a large cohort for sleep-wake
  classification with machine learning techniques. npj Digital Medicine 2(1):50

\bibitem[{Pierson et~al(2018)Pierson, Althoff, and
  Leskovec}]{pierson2018modeling}
Pierson E, Althoff T, Leskovec J (2018) Modeling individual cyclic variation in
  human behavior. In: Proceedings of the 2018 World Wide Web Conference,
  International World Wide Web Conferences Steering Committee, pp 107--116

\bibitem[{Quante et~al(2018)Quante, Kaplan, Cailler, Rueschman, Wang, Weng,
  Taveras, and Redline}]{quante18}
Quante M, Kaplan ER, Cailler M, Rueschman M, Wang R, Weng J, Taveras EM,
  Redline S (2018) Actigraphy-based sleep estimation in adolescents and adults:
  a comparison with polysomnography using two scoring algorithms. Nature and
  science of sleep 10:13

\bibitem[{Redline et~al(2014)Redline, Sotres-Alvarez, Loredo, Hall, Patel,
  Ramos, Shah, Ries, Arens, Barnhart et~al}]{redline2014sleep}
Redline S, Sotres-Alvarez D, Loredo J, Hall M, Patel SR, Ramos A, Shah N, Ries
  A, Arens R, Barnhart J, et~al (2014) Sleep-disordered breathing in
  hispanic/latino individuals of diverse backgrounds. the hispanic community
  health study/study of latinos. American journal of respiratory and critical
  care medicine 189(3):335--344

\bibitem[{Sathyanarayana(2017)}]{sathyanarayana2017}
Sathyanarayana A (2017) Computational sleep science: Machine learning for the
  detection, diagnosis, and treatment of sleep problems from wearable device
  data. PhD thesis, University of Minnesota

\bibitem[{Smith et~al(2018)Smith, McCrae, Cheung, Martin, Harrod, Heald, and
  Carden}]{smith2018use}
Smith MT, McCrae CS, Cheung J, Martin JL, Harrod CG, Heald JL, Carden KA (2018)
  Use of actigraphy for the evaluation of sleep disorders and circadian rhythm
  sleep-wake disorders: an american academy of sleep medicine clinical practice
  guideline. Journal of Clinical Sleep Medicine 14(07):1231--1237

\bibitem[{Srivastava et~al(2015)Srivastava, Mansimov, and
  Salakhudinov}]{composite_lstm_autoencoder}
Srivastava N, Mansimov E, Salakhudinov R (2015) Unsupervised learning of video
  representations using lstms. In: International conference on machine
  learning, pp 843--852

\bibitem[{Swan(2009)}]{swan2009qs}
Swan M (2009) Emerging patient-driven health care models: an examination of
  health social networks, consumer personalized medicine and quantified
  self-tracking. International journal of environmental research and public
  health 6(2):492--525

\bibitem[{Taheri(2006)}]{taheri06}
Taheri S (2006) The link between short sleep duration and obesity: we should
  recommend more sleep to prevent obesity. Archives of disease in childhood
  91(11):881--884

\bibitem[{Weiss et~al(2010)Weiss, Johnson, Berger, and Redline}]{weiss10}
Weiss AR, Johnson NL, Berger NA, Redline S (2010) Validity of activity-based
  devices to estimate sleep. Journal of Clinical Sleep Medicine 6(04):336--342

\bibitem[{Wong et~al(2011)Wong, Colley, Gorber, and Tremblay}]{wong2011actical}
Wong SL, Colley R, Gorber SC, Tremblay M (2011) Actical accelerometer sedentary
  activity thresholds for adults. Journal of Physical Activity and Health
  8(4):587--591

\bibitem[{Wright~Jr et~al(2020)Wright~Jr, Linton, Withrow, Casiraghi, Lanza,
  de~la Iglesia, Vetter, and Depner}]{wright2020sleep}
Wright~Jr KP, Linton SK, Withrow D, Casiraghi L, Lanza SM, de~la Iglesia H,
  Vetter C, Depner CM (2020) Sleep in university students prior to and during
  covid-19 stay-at-home orders. Current Biology

\bibitem[{Yang et~al(2012)Yang, Ho, Chen, and Chien}]{yang2012exercise}
Yang PY, Ho KH, Chen HC, Chien MY (2012) Exercise training improves sleep
  quality in middle-aged and older adults with sleep problems: a systematic
  review. Journal of physiotherapy 58(3):157--163

\bibitem[{Zhang et~al(2018)Zhang, Cui, Mueller, Tao, Kim, Rueschman, Mariani,
  Mobley, and Redline}]{zhang18}
Zhang GQ, Cui L, Mueller R, Tao S, Kim M, Rueschman M, Mariani S, Mobley D,
  Redline S (2018) The national sleep research resource: towards a sleep data
  commons. Journal of the American Medical Informatics Association

\end{thebibliography}
